\documentclass[letterpaper, 10 pt, conference]{ieeeconf}
\IEEEoverridecommandlockouts            	
                                                          
\usepackage{graphics} 	
\usepackage{epsfig} 	
\usepackage{mathptmx} 	
\usepackage{times} 		
\usepackage{booktabs}

\usepackage{cite}
\usepackage{amsmath,amssymb,amsfonts}
\usepackage{algorithm,algcompatible}
\usepackage{algpseudocode}
\usepackage{graphicx}
\usepackage{textcomp}
\usepackage{xcolor}
\usepackage[hidelinks]{hyperref}
\usepackage{caption}
\usepackage{subcaption}
\usepackage{enumerate}
\usepackage{multirow}
\usepackage{tabularray}

\renewcommand{\COMMENT}[2][.5\linewidth]{%
  \leavevmode\hfill\makebox[#1][l]{$\triangleright$~#2}}

\title{\LARGE \bf
Robot Agnostic Visual Servoing considering kinematic constraints enabled by a decoupled network trajectory planner structure 
}

\author{Constantin Schempp and Christian Friedrich$^{1}$
\thanks{$^{1}$ All authors are with the Department of Mechanical Engineering and Mechatronics,
Karlsruhe University of Applied Sciences (HKA), Germany,
{\tt\small \{constantin.schempp, christian.friedrich\}@h-ka.de}}%
}

\begin{document}

\maketitle
\thispagestyle{empty}
\pagestyle{empty}

\begin{abstract}
We propose a visual servoing method consisting of a detection network and a velocity trajectory planner.
First, the detection network estimates the objects position and orientation in the image space.
Furthermore, these are normalized and filtered. The direction and orientation is then the input to the trajectory planner,
which considers the kinematic constrains of the used robotic system.
This allows safe and stable control, since the kinematic boundary values are taken into account in planning.
Also, by having direction estimation and velocity planner separated, the learning part of the method does not directly influence the control value.
This also enables the transfer of the method to different robotic systems without retraining, therefore being robot agnostic.
We evaluate our method on different visual servoing tasks with and without clutter on two different robotic systems.
Our method achieved mean absolute position errors of $<0.5$ mm and orientation errors of $<1$°. Additionally, we transferred the method to a new 
system which differs in robot and camera, emphasizing robot agnostic capability of our method.
\end{abstract}

\section{Introduction}
Visual Servoing (VS) is used to control the motion of a robot based on visual information, like image input from a camera.
The camera providing the image stream can be attached to the robot (eye-in-hand), or observe the robot (eye-to-hand).
Many robot guided manipulation tasks use VS for vision-guided motion to enable grasping, inserting or pushing.

VS can be divided into image-based visual servo (IBVS) and position-based visual servo (PBVS)\cite{Chaumette_2020, kragic2002survey}.
IBVS requires visual features extracted from an image to design the control-law.
These features are often hand-crafted and mainly influence the servoing behavior \cite{nicolas_andreff, emalis}.
The control-law is then used to minimize the error between desired and current features, thus servoing the robot to the desired pose.
PBVS uses the pose of the camera relative to the object to establish an error model.
Here, the pose of the object is estimated through image characteristics, thus not directly using information about the objects geometry.
Various methods exist for object pose estimation, such as \cite{Hu_2020_CVPR, Song_2020_CVPR}.

Recent advancement in deep learning utilize CNNs to directly learn a visual controller for a specific task \cite{Yan_Tao_Xu_2022, Al-Shanoon_Lang_2022, pmlr-v155-katara21a}.
Instead of handcrafted features, these methods require large amount of training data from real experiments and camera pose estimation based on input images.

The designed control law in all approaches is usually not considering the kinematic constrains of the robot used.
Thus, learning-based approaches directly influence the servoing behavior without ensuring that kinematic limits are not violated.
Furthermore, the trajectory profile cannot be adapted for specific tasks.

To achieve an end-to-end VS method, considering kinematic constrains for robot agnostic servoing tasks, we propose a
hybrid control method combining the direction estimation from a detection network with a velocity trajectory planner.
Because the detection network only estimates the target direction, it has no influence on the velocity profile.
This decoupling makes it possible to deploy different robot systems for VS tasks without retraining.
Furthermore, it is ensured that robot motion does not violate any kinematic constraints, as kinematic boundary values are taken into account in planning.

The main contributions of this paper are:
\begin{itemize}
\item A robust and stable visual servoing method, considering kinematic constraints
\item Task adaptable trajectory profiles
\item Transferability of the method to different robotic systems without retraining
\end{itemize}

\begin{figure}[t!]
\centering
\includegraphics[scale=1.5]{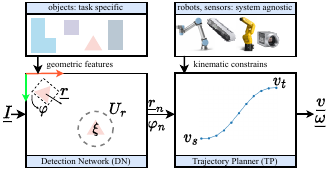}
\caption{Overview of the method. The task specific objects define the detection network used.
The decoupled trajectory planner is parameterized by the kinematic constrains of the robot system.}
\label{fig:method}
\end{figure}

\section{Related Work}
Traditional IBVS uses image features to establish the interaction matrix used for controlling the robot \cite{Hutchinson_Hager_Corke_1996}.
Since manually defining these features is tedious and not applicable in a more general way,
approaches exist that try to automatically extract these features \cite{Costanzo_De_Maria_Natale_Russo_2024, Luo_Chen_Quan_Zhang_Liu_2020}.
Also, specialized features are designed to solve the servoing task more precisely.
For example \cite{Yan_Tao_Xu_2022} extracts line- and point-features using a segmentation network to construct the interaction matrices.
End-to-end methods utilize sophisticated deep learning methods to directly estimate the velocity or target pose to steer the robot \cite{Bateux_Marchand_Leitner_Chaumette_Corke_2017, Wu_Jin_Liu_Yu_Yang_2022, Tokuda_Arai_Kosuge_2021}.
For example KOVIS \cite{Puang_Peng_Tee_Jing_2020} is used in manipulation tasks where the velocity is estimated based on predicted keypoints from a keypoint extraction network.
Coarse-to-Fine Visual Servoing (CFVS)\cite{Lu_Chen_Lee_Hsu_2023} is used in peg-in-hole insertion tasks and utilizes visual servoing by an offset prediction network, which estimates the pose offset from pointcloud data.
For considering kinematic constraints, trajectory planning is used to guarantee motion profiles within the kinematic limits of the robot \cite{Biagiotti_Melchiorri_2008}.
Here, polynomial cubic, quintic or trigonometric trajectories are commonly used.
By all approaches, IBVS and learning based methods, the robot is controlled directly by the estimated velocity.
Hereby, especially the learning based methods, the network directly influences the servoing behavior of the robot without considering the kinematic constrains of the robot system used.
Therefore, it is not possible to transfer these methods to robot-sensor-configurations that were not involved in the learning process, since no stable execution can be guaranteed.
That is why we propose a method which separates the direction estimation and the underlying control law. 
By changing the robot system used, the trajectory planner can be parameterized to ensure stable control without the need to
retrain the detection network.

\section{Method}
In this work, we present a hybrid Visual Servoing structure, consisting of a detection network DN and a velocity trajectory planner TP.
An overview of the method is given in \autoref{fig:method}. 
For a better overview, we have listed the nomenclature used in this work in \autoref{tab:nomenclature}.

The detection network is used to learn the geometric features describing the object relevant to the task.
Using the features, the oriented bounding box (OBB) is estimated, from which both the necessary translational and rotational directions to the target in the image area can be derived. The DN is task specific and can be selected independently to the trajectory planner.
Since there are established models for such problems in the deep learning area, pretrained models can be adapted to suit the specific task. 

The velocity trajectory planner utilizes the estimated directions to plan a velocity profile, which considers the kinematic constrains of the robot system used.
The planner is preceded by a filter to smooth detection results to ensure robustness and consider detection errors.

\begin{table}[t]
\caption{Nomenclature used in this work. Underlined lower case symbols represent vectors, underlined upper case matrices. Others are scalars.}
\label{tab:nomenclature}
\begin{center}
\begin{tabular}{lll}
\hline
\textbf{Symbol} & \textbf{Unit} & \textbf{Meaning} \\ \hline
DN & - & Detection Network \\
TP & - & Trajectory Planner \\
$T_D$ & [s] & cycle time detection network \\
$T_R$ & [s] & cycle time robot controller \\
$\underline{I}$ & [px, px] & Image of size $w,h$ \\
$\xi$ & [px, px] & target position in image \\
$c$ & [px, px] & bounding box center \\
$\underline{r}$, $\varphi$ & [px, px], [rad] & direction, orientation \\
$\underline{r}_n$, $\varphi_n$ & -, - & norm. direction, orientation \\
$\underline{r}_{n,f}$, $\varphi_{n,f}$ & -, - & filtered direction, orientation \\
$U_r$ & [px] & vicinity of direction \\
$U_{\varphi}$ & [rad] & vicinity of orientation \\
$N$ & - & filter size \\
$K$ & - & number of trajectory points \\
$v_s$, $\dot{v}_s$, $\ddot{v}_s$ & [$\frac{m}{s}$], [$\frac{m}{s^2}$], [$\frac{m}{s^3}$] & start values \\
$v_t$, $\dot{v}_t$ , $\ddot{v}_t$ & [$\frac{m}{s}$], [$\frac{m}{s^2}$], [$\frac{m}{s^3}$] & target values \\
$t_s$, $t_t$ & [s], [s] & start time, target time \\
$v_{max}$, $\dot{v}_{max}$, $\ddot{v}_{max}$ & [$\frac{m}{s}$], [$\frac{m}{s^2}$], [$\frac{m}{s^3}$] & translational constraints \\
$\omega_{max}$, $\dot{\omega}_{max}$, $\ddot{\omega}_{max}$ & [$\frac{rad}{s}$], [$\frac{rad}{s^2}$], [$\frac{rad}{s^3}$] & radial constraints \\
$\underline{v}$, $\underline{\omega}$ & [$\frac{m}{s}$], [$\frac{rad}{s}$]  & velocity trajectory \\
\hline
\end{tabular}
\end{center}
\end{table}

\subsection{Detection Network}
We first train the detection network to predict the OBB of the desired object.
The detection network takes a image $I_{w,h}$ of width $w$ and height $h$ as input and predicts the oriented bounding box (OBB) of the object.
The OBB is defined by the center $c=[x, y]$, width $w$, height $h$ and orientation $\varphi$ of the bounding box:
\begin{align}
\text{OBB} = [x, y, w, h, \varphi] .
\end{align}
From this, we calculate the direction $r$ to the target $\xi$ by $\underline{r}=\xi - c$. Furthermore, direction $\underline{r}$ and orientation $\varphi$ of the OBB are
normalized, addressing the problem of deceleration:

\begin{align}
\label{eq:r_norm}
\underline{r}_n &=
\begin{cases}
    \frac{\underline{r}}{\Vert \underline{r} \Vert}, &\text{if } \Vert \underline{r} \Vert \geq U_r\\
    \frac{\underline{r}}{U_r}, &\text{if } \Vert \underline{r} \Vert < U_r\\
    0, &\text{if aligned} 
\end{cases} \\
\varphi_n &=
\begin{cases}
    \frac{\varphi}{\Vert \varphi \Vert},& \text{if } \Vert \varphi \Vert \geq U_{\varphi}\\
    \frac{\varphi}{U_{\varphi}},& \text{if } \Vert \varphi \Vert < U_{\varphi}\\
    0,              & \text{if aligned}
\end{cases}
\label{eq:phi_norm}
\end{align}
Here, $U$ is the vicinity at which the normalized direction and orientation is downscaled relative to the target.
This ensures high velocities when further away and a damped approach near the destination.
The time required to capture an image and make a prediction is described by $T_D$ and is called the cycle time of the DN.

\subsection{Velocity Trajectory Planner}
For the sake of simplicity, we plan in Cartesian space. As constraints, we take the speed and acceleration limits of the EEF in Cartesian space, as specified by the robot manufacturer. It is also possible to perform trajectory planning in joint space using the Jacobian.
The trajectory planner consists of a preceding filter, smoothing the detection results to counteract outliers and noise.
A moving average (MAVG) filter of size $N$ is used to filter the normalized direction $\underline{r}_n$ and orientation $\varphi_n$:
\begin{align}
\label{eq:normalize_r}
\underline{r}_{n,f} &= \sum_{i=1}^N N^{-1} \cdot \underline{r}_{n, i} \\
\varphi_{n,f} &= \sum_{i=1}^N N^{-1} \cdot \varphi_{n, i} \label{eq:normalize_phi} 
\end{align}

Next, we define each a function mapping the filtered direction and orientation to velocity, acceleration and jerk:
\begin{align}
f_r(\underline{r}_{n,f}) &= v  &g_r(\underline{r}_{n,f}) &= \dot{v}  &h_r(\underline{r}_{n,f}) &= \ddot{v} \\
f_{\varphi}(\varphi_{n,f}) &= \omega  &g_{\varphi}(\varphi_{n,f}) &= \dot{\omega}  &h_{\varphi}(\varphi_{n,f}) &= \ddot{\omega}
\end{align}
In this work we use
\begin{align} \label{eq:mapping_r}
f_r &= v_{max} \cdot \underline{r}_{n,f} & g_r&=\dot{v}_{max} \cdot \underline{r}_{n,f}  & h_r&=\ddot{v}_{max} \cdot \underline{r}_{n,f}\\
f_{\varphi} &= \omega_{max} \cdot \varphi_{n,f} & g_{\varphi} &= \dot{\omega}_{max} \cdot \varphi_{n,f}  & h_{\varphi} &= \ddot{\omega}_{max} \cdot \varphi_{n,f} \label{eq:mapping_phi}
\end{align}
Given that direction and orientation are normalized by \eqref{eq:normalize_r} and \eqref{eq:normalize_phi}, the calculated values from \eqref{eq:mapping_r} and \eqref{eq:mapping_phi} cannot exceed the velocity and acceleration constraints.
Furthermore, we define start and target values for velocity, acceleration and jerk at start and end time.
For better overview, only translational velocity is mentioned here. However, the same applies to the radial velocity:
\begin{align}
v(t_s) &= v_s   &\dot{v}(t_s)&=\dot{v}_s &\ddot{v}(t_s)&=\ddot{v}_s \\
v(t_t) &= v_t   &\dot{v}(t_t)&=\dot{v}_t  &\ddot{v}(t_t)&=\ddot{v}_t
\end{align}
To fit the start and target conditions, we use a quintic polynomial, written as a linear equation system (LES):
\begin{align}
\underbrace{\begin{bmatrix}
           v_s \\
           \dot{v}_s \\
           \ddot{v}_s \\
           v_t \\
           \dot{v}_t \\
           \ddot{v}_t
\end{bmatrix}}_\text{\underline{b}} =
\underbrace{\begin{bmatrix}
1 & t_s & t_s^2 & t_s^3 & t_s^4 & t_s^5 \\
0 & 1 & 2t_s & 3t_s^2 & 4t_s^3 & 5t_s^4 \\
0 & 0 & 2 & 6t_s & 12t_s^2 & 20t_s^3 \\
1 & t_t & t_t^2 & t_t^3 & t_t^4 & t_t^5 \\
0 & 1 & 2t_t & 3t_t^2 & 4t_t^3 & 5t_t^4 \\
0 & 0 & 2 & 6t_t & 12t_t^2 & 20t_t^3 \\
\end{bmatrix}}_\text{\underline{M}}
\underbrace{\begin{bmatrix}
a_0 \\
a_1 \\
a_2 \\
a_3 \\
a_4 \\
a_5
\end{bmatrix}}_\text{\underline{q}}
\end{align}
From this, the unknown coefficients of the polynomial can be calculated with
\begin{align}
\underline{q} = \underline{M}^{-1} \cdot \underline{b}
\end{align}

\begin{figure*}[!ht]
\begin{tabular}{cccccc}
\includegraphics[width=0.145\textwidth]{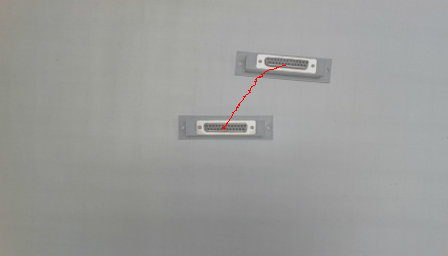} & \includegraphics[width=0.145\textwidth]{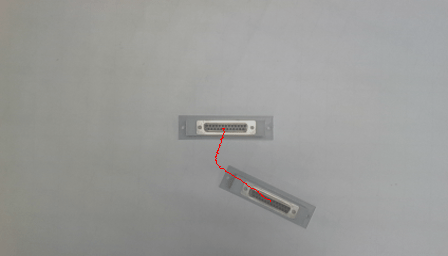} &
\includegraphics[width=0.145\textwidth]{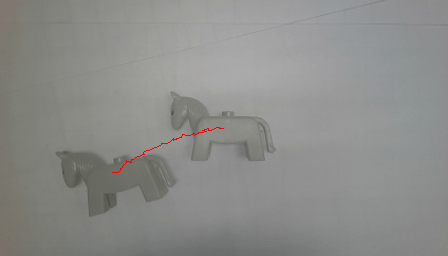} & \includegraphics[width=0.145\textwidth]{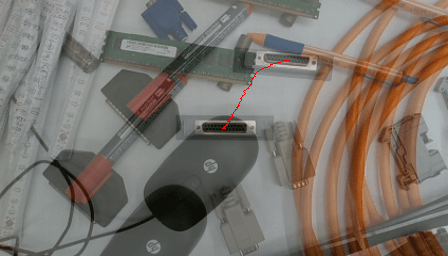} &
\includegraphics[width=0.145\textwidth]{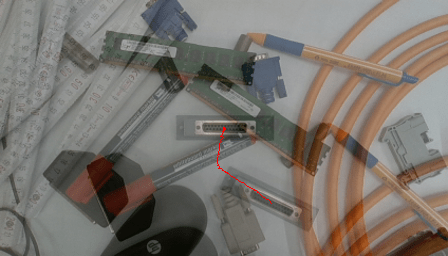} & \includegraphics[width=0.145\textwidth]{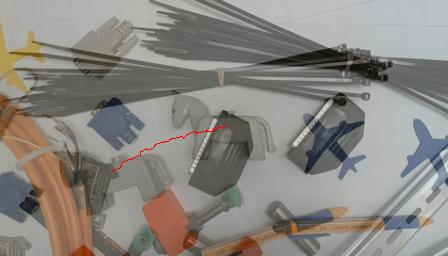} \\
\includegraphics[width=0.145\textwidth]{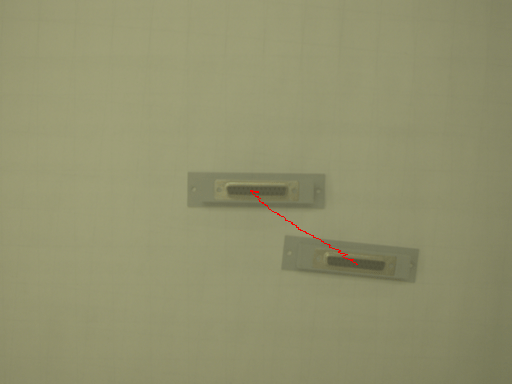} & \includegraphics[width=0.145\textwidth]{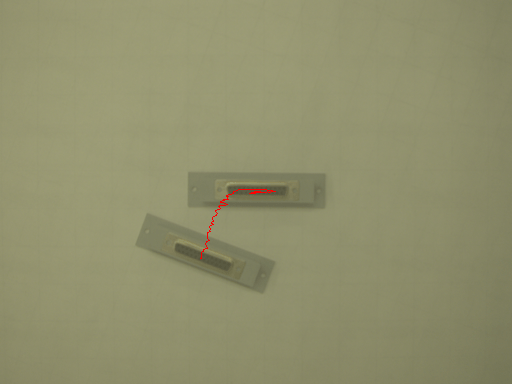} &
\includegraphics[width=0.145\textwidth]{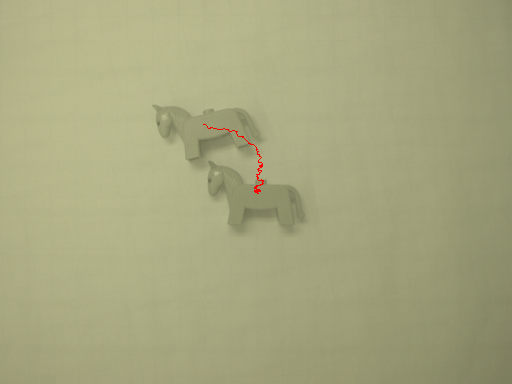} & \includegraphics[width=0.145\textwidth]{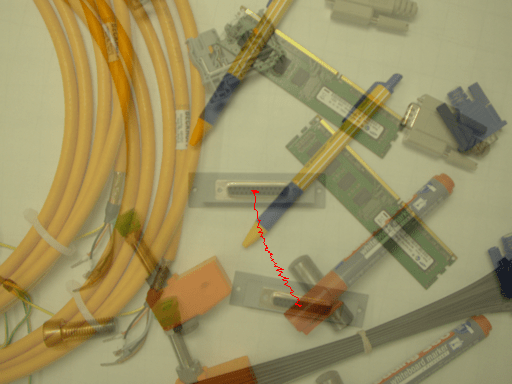} &
\includegraphics[width=0.145\textwidth]{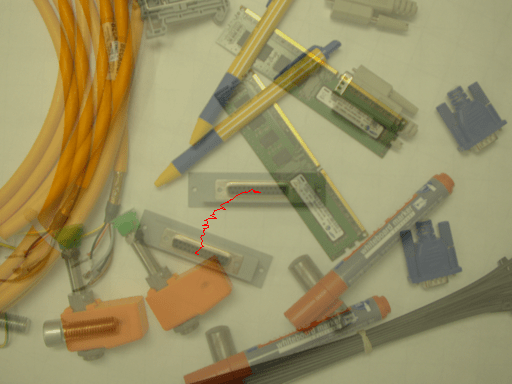} & \includegraphics[width=0.145\textwidth]{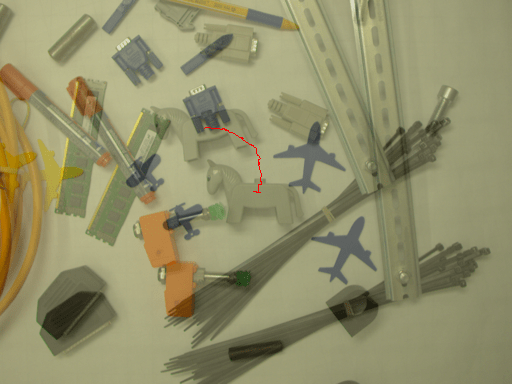} \\
\end{tabular}
\caption{Results of the visual servoing task. First row shows the task solved with the UR5e robot, second row with the LR Mate 200iD/7L.
Each column depicts a different initial alignment setting in the normal and clutter scene. The red path describes the trajectory of the detected OBB.}
\label{fig:scenes}
\end{figure*}
The cycle time $T_D$ and cycle time of the robot controller $T_R$ define the number of trajectory points $K=\frac{T_D}{T_R}$, with the condition $T_R \ll T_D$.
That is, we plan the velocity trajectory $\underline{v}$, $\underline{\omega}$ until the next prediction from the detection network is available.

The algorithms for estimating the direction and orientation and calculating the velocity trajectory are written in 
Algorithm \autoref{alg:detection_network} and Algorithm \autoref{alg:trajectory_planner} respectively.
After the first cycle, the algorithms run in parallel and the estimated $\underline{r}_n$ and $\varphi_n$ from Algorithm \autoref{alg:detection_network} are used by Algorithm \autoref{alg:trajectory_planner} to calculate $\underline{v}$ and $\underline{\omega}$.

\begin{figure}
\centering
\includegraphics[scale=1.05]{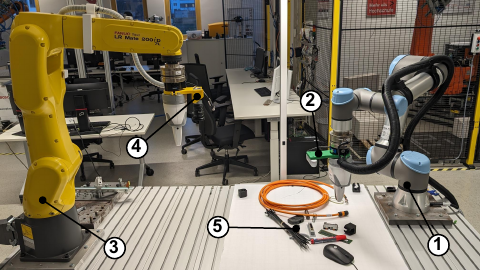}
\caption{Experimental setup of the visual servoing task, consisting of (1) UR5e robot, (2) D435 camera, (3) LR Mate 200iD/7L, (4) acA2040-35gc and (5) task setting, here clutter.}
\label{fig:setup}
\end{figure}

\section{Experiments}
We are conducting a series of experiments to measure the performance of our approach.
We are interested in precision, efficiency and transferability.
This is why we aim to determine: (1) How does our method compare to other baselines in 3-DoF visual servoing with a small and a large initial alignment error?
(2) How quickly does our method converge to the target $\xi$? And finally, (3) is it possible to transfer the method without retraining to a new robot-camera system setup
while not impairing precision and efficiency?

\begin{algorithm}[!b]
\caption{Calculation of the normalized direction and rotation during execution of the VS task.}
\label{alg:detection_network}
\begin{algorithmic}
\Require $I$	\COMMENT{image from camera stream}
\Ensure $\underline{r}_n$, $\varphi_n$

\While{true}
\State $c$, $\varphi$ $\gets \text{DN}(\underline{I})$
\State $\underline{r} \gets \xi - c$
\State $\underline{r}_n$, $\varphi_n \gets $ normalize($\underline{r}$, $\varphi$)	\COMMENT{according to \eqref{eq:r_norm}, \eqref{eq:phi_norm}}
\EndWhile
\end{algorithmic}
\end{algorithm}

\subsection{Experimental Settings}
The experiments are carried out using two different robot systems: 
a UR5e from Universal Robots with a D435 camera from Intel RealSense (60 fps) and a LR Mate 200iD/7L from Fanuc equipped with a acA2040-35gc camera from Basler (30 fps). The cameras on both systems are eye-in-hand and we control the robot motion in Cartesian space.
Inference of the detection network is carried out on a NVIDA A4000 GPU.
For interfacing all the components, detection network and trajectory planner, we use Robot Operating System (ROS)\cite{ROS} for our visual servoing pipeline.
The setup is depicted in \autoref{fig:setup}

\begin{algorithm}[!b]
\caption{Calculation of the velocity trajectory profile during execution of the VS task.}
\label{alg:trajectory_planner}
\begin{algorithmic}
\Require $\underline{r}_n$, $\varphi_n$
\Ensure $\underline{v}$, $\underline{\omega}$

\State $v_s$, $\dot{v}_s$, $\ddot{v}_s$, $\omega_s$, $\dot{\omega}_s, \ddot{\omega}_s \gets 0$
\State $t_s$, $t_t \gets 0, T_D$
\State $\underline{B}[3,N] \gets 0$ \COMMENT{buffer of size $3\times N$}

\While{true}
\State $\underline{B} \gets $push($B$, [$\underline{r}_n$, $\varphi_n$])
\State $\underline{r}_{n,f}, \varphi_{n,f} \gets N^{-1} \cdot \text{sum}(B)$ \COMMENT{calculate moving average}
\State $v_t$, $\dot{v}_t$, $\ddot{v}_t$ $\gets f_r(\underline{r}_{n,f})$, $g_r(\underline{r}_{n,f})$, $h_r(\underline{r}_{n,f})$
\State $\omega_t$, $\dot{\omega}_t$, $\ddot{\omega}_t$ $\gets f_{\varphi}(\varphi_{n,f})$, $g_{\varphi}(\varphi_{n,f})$, $h_{\varphi}(\varphi_{n,f})$
\State $\underline{q} \gets \underline{M}^{-1} \cdot \underline{b}$	\COMMENT{polynomial coefficients}
\State $\underline{t} \gets $linspace($t_s$, $t_t$, $K$) \COMMENT{evenly spaced timesteps}
\State $\underline{v}$, $\underline{\omega} \gets $evalPolynomial($\underline{q}$, $\underline{t}$)
\For{$i=0$ to $K-1$}
\State sendToRobot($\underline{v}[i]$, $\underline{\omega}[i]$)
\State waitUntil($T_R$)
\EndFor
\State $v_s$, $\dot{v}_s$, $\ddot{v}_s \gets v_t, \dot{v}_t, \ddot{v}_t$
\State $\omega_s$, $\dot{\omega}_s$, $\ddot{\omega}_s \gets \omega_t, \dot{\omega}_t, \ddot{\omega}_t$
\EndWhile
\end{algorithmic}
\end{algorithm}

\subsection{Tasks} \label{subsec:tasks}
In the experiments, the robot arm executes two different tasks: aligning a dsub connector and a toy horse in 3-DoF.
The goal of both tasks is to align with the target $\xi$, which is set as the image center.
To detect the objects in the image, we finetune a pre-trained yolov8 network \cite{Jocher_Ultralytics_YOLO_2023} for each object using transfer learning.
For training, we labeled 200 real images and split them into 75\% training and 25\% validation data.
We trained for 100 epochs with a batch size of 16 and learning rate of $0.01$.
The accuracy of the object detection of the trained network should be appropriately high since it affects the accuracy of positioning.
After finetuning, we reached an accuracy of $\text{mAP50}=0.995$ and $\text{mAP50-95}=0.724$ for the dsub connector, 
$\text{mAP50}=0.995$ and $\text{mAP50-95}=0.499$ for the toy horse. 

\begin{figure}[!t]
\centering
\includegraphics[scale=0.465]{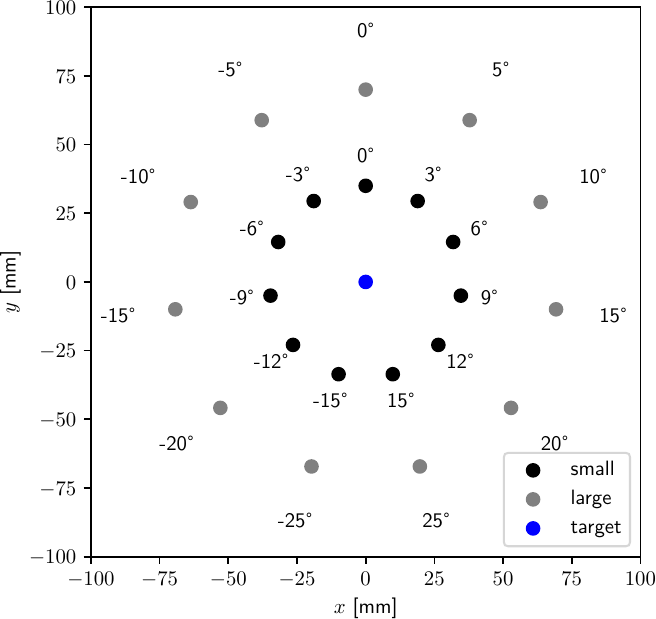}
\caption{Initial errors used for the experiments. Black indicates small initial error, gray large initial error and blue the target position.
Next to the positions, the initial rotation error is written.}
\label{fig:initial_error}
\end{figure}

We use two different scenes for our experiments. The \textbf{normal} scene has only the object of interest placed in the working area, the \textbf{clutter} scene also includes various disturbance objects.
For each scene, we consider two different sets of initial alignment errors.
Small initial alignment error: 35 mm error distance arranged circularly around the target, with rotational errors ranging from $-15^\circ$ to $15^\circ$, in $3^\circ$ steps. 
Large initial alignment error: 70 mm error distance arranged circularly around the target, with rotational errors ranging from $-25^\circ$ to $25^\circ$, in $5^\circ$ steps. 
The distribution of the initial errors is depicted in \autoref{fig:initial_error}.

\begin{figure*}[!t]
\begin{tabular}{cccccc}
\includegraphics[width=0.145\textwidth]{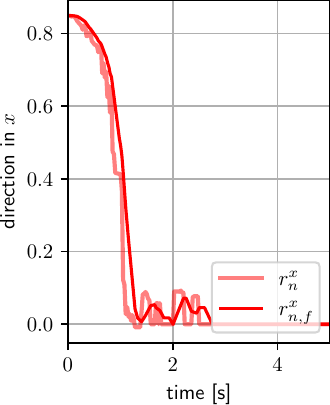} & 
\includegraphics[width=0.145\textwidth]{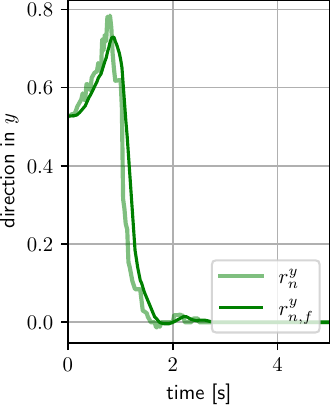} &
\includegraphics[width=0.145\textwidth]{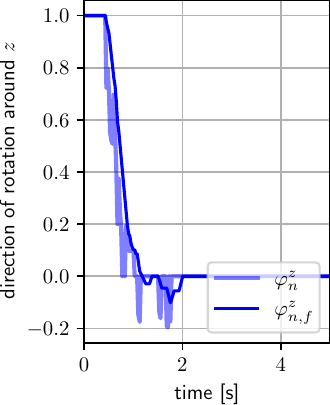} & 
\includegraphics[width=0.145\textwidth]{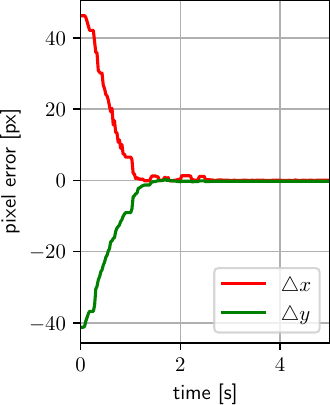} & 
\includegraphics[width=0.145\textwidth]{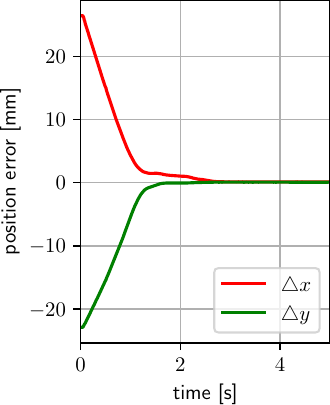} &  
\includegraphics[width=0.145\textwidth]{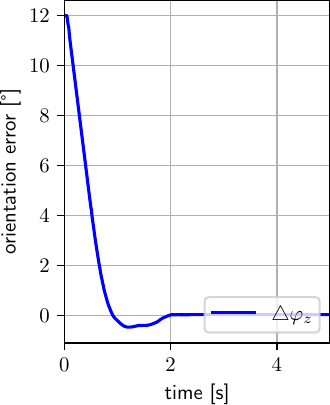} \\
(a) & (b) & (c) & (d) & (e) & (f)
\end{tabular}
\caption{Progression of the dsub connector visual servoing task using the UR5e with large initial error. (a) detected normalized $x$-direction, (b) detected normalized $y$-direction, 
(c) detected normalized orientation, (d) pixel error between OBB center and target $\xi$, (e) position error of the robot end effector and (f) the orientation error of the robot end effector.}
\label{fig:results}
\end{figure*}

\subsection{Evaluation Metrics}
We use the mean absolute error to measure the precision of our method.
For efficiency, we measure the time to reach a translational error threshold of 1 px in the image space and rotational error threshold of $1^\circ$.
The transferability is determined by using the method on two completely different robot systems and looking at the deviation in precision and efficiency.
Each task is conducted for 20 s.

\subsection{Baselines \& Results} \label{subsec:baselines}
As baselines, we compare our approach to methods using visual servoing in alignment tasks.
The two baselines used are briefly described in the following section.

Work \cite{Ma_Liu_Zhang_Xu_Zhang_Wu_2020}: A robotic assembly system for small components precision assembly. The robot conducts two tasks, grasping and pose alignment.
For alignment, image features and a IBVS control method is deployed.

Work \cite{Yan_Tao_Xu_2022}: A deep neural network is combined with feature pyramid network structure. This is used to extract point and line features from the image.
The interaction matrices are constructed based on these features.
\autoref{tab:results} and \autoref{tab:results_horse} depict the mean absolute error for the normal and clutter scene for each task and robot setup.
Additionally, the mean time to reach the goal for each degree of freedom is mentioned.
\autoref{fig:scenes} shows a selection of the initial and final state of the servoing for the normal and clutter scene and \autoref{fig:results} the progression of the VS task.

\begin{table}[t]
\caption{Results of the visual servoing experiments for task \textbf{dsub}. For every robot system, the mean absolute errors are depicted.
n stands for the normal scene and c for the clutter scene.}
\label{tab:results}
\begin{center}
\begin{tabular}{lllllllll}
\hline
robot                                                    & \multicolumn{4}{c}{UR5e}                                                                          & \multicolumn{4}{c}{LR Mate 200iD/7L}                                                          \\
camera                                                   & \multicolumn{4}{c}{D435}                                                                          & \multicolumn{4}{c}{acA2040-35gc}                                                              \\ \hline
\begin{tabular}[c]{@{}l@{}}initial \\ error\end{tabular} & \multicolumn{2}{c}{small}                     & \multicolumn{2}{c}{large}                         & \multicolumn{2}{c}{small}                     & \multicolumn{2}{c}{large}                     \\
scene                                                    & \multicolumn{1}{c}{n} & \multicolumn{1}{c}{c} & \multicolumn{1}{c}{n} & \multicolumn{1}{c|}{c}    & \multicolumn{1}{c}{n} & \multicolumn{1}{c}{c} & \multicolumn{1}{c}{n} & \multicolumn{1}{c}{c} \\ \hline
$\bigtriangleup x$ {[}mm{]}                              & 0.25                  & 0.18                  & 0.15                  & \multicolumn{1}{l|}{0.18} & 0.25                  & 0.42                  & 0.21                  & 0.40                  \\
$\bigtriangleup y$ {[}mm{]}                              & 0.10                  & 0.05                  & 0.21                  & \multicolumn{1}{l|}{0.05} & 0.08                  & 0.11                  & 0.13                  & 0.09                  \\
$\bigtriangleup \varphi_z$ {[}°{]}                       & 0.10                  & 0.07                  & 0.09                  & \multicolumn{1}{l|}{0.07} & 0.12                  & 0.67                  & 0.14                  & 0.44                  \\
$t_r$ [s]				  & 1.36				  &	1.30		  & 2.49				 & \multicolumn{1}{l|}{2.49}				&  5.28 & 4.73 & 6.29 & 6.55 \\
$t_{\varphi}$ {[}s{]}                                    & 0.59                  & 0.53                  & 0.93                  & \multicolumn{1}{l|}{0.95} & 3.17                  & 3.87                  & 3.39                  & 3.77                  \\ \hline
\end{tabular}
\end{center}
\vspace{-5.5mm}
\end{table}
 
\begin{figure*}[t]
\begin{tabular}{cccc}
\includegraphics[width=0.23\textwidth]{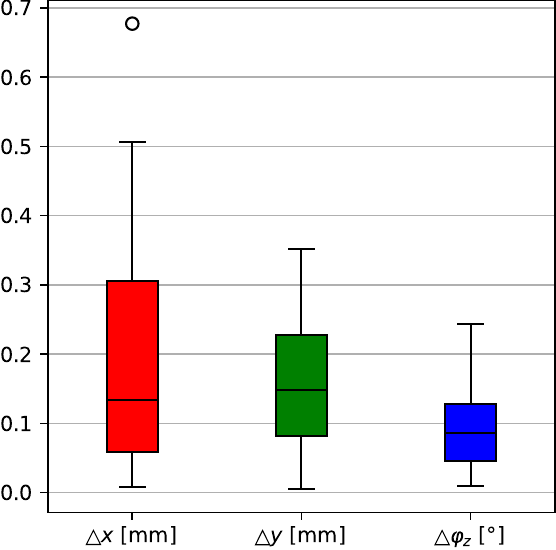} & 
\includegraphics[width=0.23\textwidth]{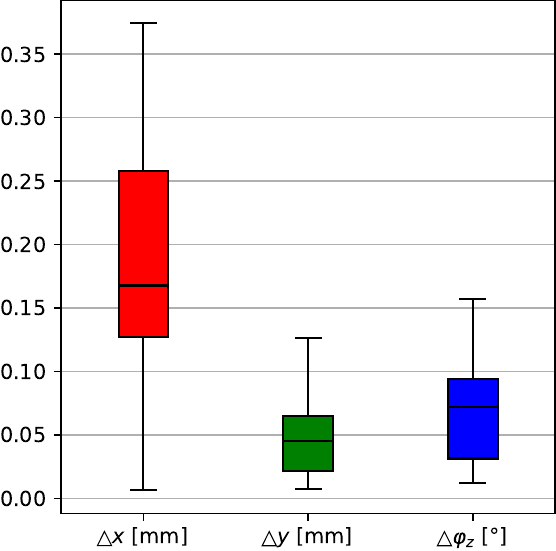} &
\includegraphics[width=0.23\textwidth]{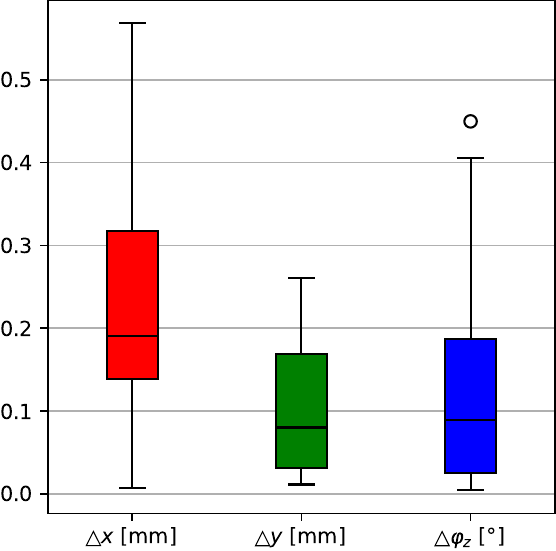} & 
\includegraphics[width=0.23\textwidth]{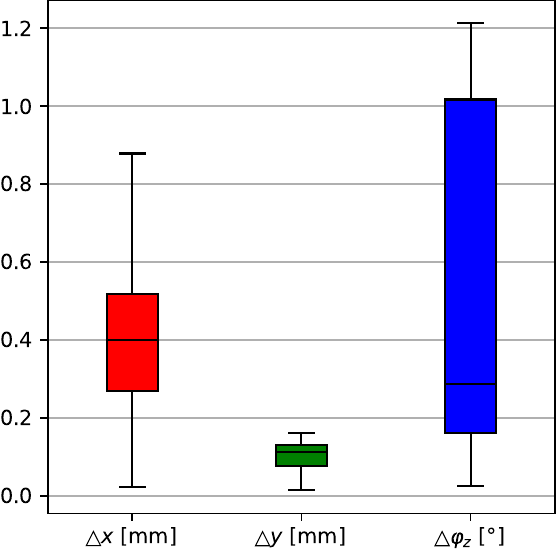} \\
(a) & (b) & (c) & (d)
\end{tabular}
\caption{Boxplots of the position and orientation errors for the dsub connector. (a) UR5e normal scene, (b) UR5e clutter scene, (c) Fanuc normal scene and (d) Fanuc clutter scene.}
\label{fig:boxplots}
\end{figure*}

\textbf{Precision.} 
The position errors in the $x$ and $y$ direction and orientation errors are listed for each initial alignment error.
The low errors of $<0.5$ mm in position and $<1.0$ ° in orientation for the dsub connector show, that our method is able to precisely align the object.
The errors of the toy horse object of $<1.1$ mm and $<1.0$ ° are higher, because the precision of the corresponding detection network is lower, see \autoref{subsec:tasks}.
Furthermore, \autoref{tab:benchmark} shows that our method is able to achieve equal or better precision than the baselines methods.
Here, we additionally listed the relative remaining error after aligning as a percentage to obtain an evaluation independent of the initial error.
\autoref{fig:boxplots} illustrates the distribution of errors and outliers.

\textbf{Efficiency.}
Using the UR5e setup, our method aligned to the target in about 1.3 s with a small initial error and in about 2.5 s with the large initial error.
Using the Fanuc setup, we had to decrease the velocity constrains to prevent overshoot, because the used camera system has a lower frame rate, thus limiting the reaction time.
It should be noted that the alignment time depends on the kinematic constrains of the robot-sensor-configuration used. 
Our approach is able to align the desired object in a reasonable amount of time while being faster than the baseline methods.
In addition to the time to reach the target, the average speed is also listed in mm/s and rad/s.

\begin{figure}[t]
\begin{tabular}{cc}
\includegraphics[scale=0.58]{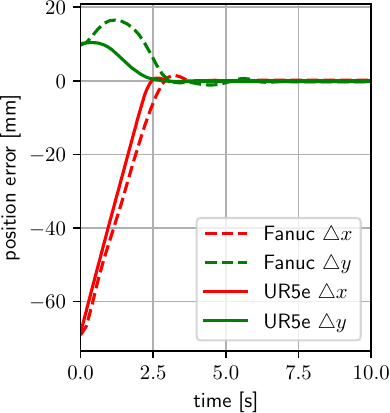} & 
\includegraphics[scale=0.58]{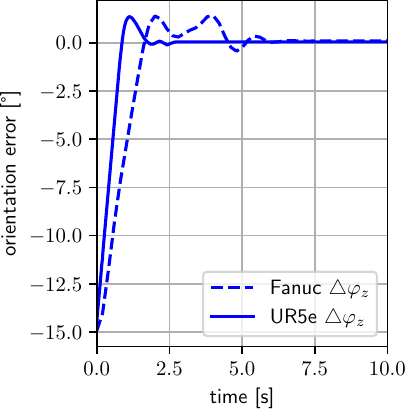}
\end{tabular}
\caption{Comparison of position and orientation error profiles of both systems for the same task (dsub, large initial error, 15°). Dashed lines represent LR Mate 200iD/7L, solid line UR5e.}
\label{fig:bothrobots}
\end{figure}

\textbf{Transferability.}
We were able to transfer our method to a system consisting of a different robot and camera.
As shown in \autoref{tab:results} and \autoref{tab:results_horse}, completing the servoing task without significant loss of precision or efficiency is possible.
We can conclude from this that our method has robot agnostic capability.
\autoref{fig:bothrobots} shows the error profiles of both systems for the same initial error.

\begin{table}[t]
\caption{Results of the visual servoing experiments for task \textbf{toy horse}. For every robot system, the mean absolute errors are depicted.
n stands for the normal scene and c for the clutter scene.}
\label{tab:results_horse}
\begin{center}
\begin{tabular}{lllllllll}
\hline
robot                                                   & \multicolumn{4}{c}{UR5e}                                                                          & \multicolumn{4}{c}{LR Mate 200iD/7L}                                                          \\
camera                                                  & \multicolumn{4}{c}{D435}                                                                          & \multicolumn{4}{c}{acA2040-35gc}                                                              \\ \hline
\begin{tabular}[c]{@{}l@{}}initial\\ error\end{tabular} & \multicolumn{2}{c}{small}                     & \multicolumn{2}{c}{large}                         & \multicolumn{2}{c}{small}                     & \multicolumn{2}{c}{large}                     \\
scene                                                   & \multicolumn{1}{c}{n} & \multicolumn{1}{c}{c} & \multicolumn{1}{c}{n} & \multicolumn{1}{c|}{c}    & \multicolumn{1}{c}{n} & \multicolumn{1}{c}{c} & \multicolumn{1}{c}{n} & \multicolumn{1}{c}{c} \\ \hline
$\bigtriangleup x$ {[}mm{]}                             & 0.98                  & 1.07                  & 1.11                  & \multicolumn{1}{l|}{0.81} & 0.78                  & 0.81                  & 0.63                  & 1.08                  \\
$\bigtriangleup y$ {[}mm{]}                             & 0.92                  & 0.36                  & 0.91                  & \multicolumn{1}{l|}{0.29} & 0.18                  & 0.34                  & 0.24                  & 0.27                  \\
$\bigtriangleup \varphi_z$ {[}°{]}                      & 0.94                  & 0.43                  & 0.94                  & \multicolumn{1}{l|}{0.32} & 0.53                  & 0.34                  & 0.49                  & 0.37                  \\
$t_r$ {[}s{]}                                           & 2.11                  & 3.41                  & 3.89                  & \multicolumn{1}{l|}{4.64} & 8.80                  & 8.22                  & 9.91                  & 10.0                  \\
$t_{\varphi}$ {[}s{]}                                   & 1.27                  & 1.74                  & 2.04                  & \multicolumn{1}{l|}{2.48} & 4.59                  & 3.64                  & 4.09                  & 4.06                  \\ \hline
\end{tabular}
\end{center}
\end{table}

\begin{table}[!t]
\caption{Comparison of the servoing results with the baselines mentioned in \autoref{subsec:baselines}. The best results are marked in bold.
Values that are not mentioned in the baselines are marked with -.}
\label{tab:benchmark}
\begin{center}
\begin{tabular}{@{}lllllll@{}}
\toprule
method                         & \multicolumn{2}{c}{\cite{Ma_Liu_Zhang_Xu_Zhang_Wu_2020}} & \multicolumn{2}{c}{\cite{Yan_Tao_Xu_2022}} & \multicolumn{2}{c}{ours} \\ \midrule
$\bigtriangleup x$ [mm], [\%]        & \textbf{0.02}                  & 0.13                  & 0.17                & 2.90             & 0.18           & 0.40  \\
$\bigtriangleup y$ [mm], [\%]       & \textbf{0.02}                  & 0.10\                 & 0.16                & 8.10               & 0.05           & 0.10  \\
$\bigtriangleup \varphi_z$ [°] & -                              & -                       & 0.18                & 10.7               & \textbf{0.07}  & 0.70  \\
$t_r$ [s], [$\frac{mm}{s}$]                      & 15                             & 1.64                    & -                   & -                    & \textbf{1.30}  & 53.8   \\
$t_{\varphi}$ [s], [$\frac{rad}{s}$]              & -                              & -                       & -                   & -                    & \textbf{0.53}  & 0.33   \\ \bottomrule
\end{tabular}
\end{center}
\end{table}

\section{Conclusion}
In this work, a system consisting of a decoupled detection network and velocity trajectory planner was proposed.
We were able to achieve precise alignment in a comparable short time, visible by position errors below one millimeter and orientation errors below one degree.
Additionally, our method can be transferred to new robot systems, making it robot agnostic. 
We showed this by performing the same experiments on two completely different systems with nearly
the same results in precision and efficiency. Because we are considering the kinematic constrains of the system used, stable and robust visual servoing is possible.
Future work will explore the possibility to extend this method to 6-DoF by using a detection network estimating point cloud OBBs.
Furthermore, the automated selection of kinematic constrains depending on the inference time and the robot dynamics will be addressed.
Finally planning of trajectories for task-dependent paths or time-optimized sequences will be considered in the future.


\bibliography{IEEEabrv, mybibfile.bib}

\begin{thebibliography}{10}
\providecommand{\url}[1]{#1}
\csname url@rmstyle\endcsname
\providecommand{\newblock}{\relax}
\providecommand{\bibinfo}[2]{#2}
\providecommand\BIBentrySTDinterwordspacing{\spaceskip=0pt\relax}
\providecommand\BIBentryALTinterwordstretchfactor{4}
\providecommand\BIBentryALTinterwordspacing{\spaceskip=\fontdimen2\font plus
\BIBentryALTinterwordstretchfactor\fontdimen3\font minus
  \fontdimen4\font\relax}
\providecommand\BIBforeignlanguage[2]{{%
\expandafter\ifx\csname l@#1\endcsname\relax
\typeout{** WARNING: IEEEtran.bst: No hyphenation pattern has been}%
\typeout{** loaded for the language `#1'. Using the pattern for}%
\typeout{** the default language instead.}%
\else
\language=\csname l@#1\endcsname
\fi
#2}}

\bibitem{Chaumette_2020}
\BIBentryALTinterwordspacing
F.~Chaumette, \emph{\BIBforeignlanguage{en}{Visual Servoing}}.\hskip 1em plus
  0.5em minus 0.4em\relax Berlin, Heidelberg: Springer Berlin Heidelberg, 2020,
  p. 1–9. [Online]. Available:
  \url{http://link.springer.com/10.1007/978-3-642-41610-1_104-1}
\BIBentrySTDinterwordspacing

\bibitem{kragic2002survey}
D.~Kragic, H.~I. Christensen, \emph{et~al.}, ``Survey on visual servoing for
  manipulation,'' \emph{Computational Vision and Active Perception Laboratory,
  Fiskartorpsv}, vol.~15, p. 2002, 2002.

\bibitem{nicolas_andreff}
\BIBentryALTinterwordspacing
N.~Andreff, B.~Espiau, and R.~Horaud, ``Visual servoing from lines,'' \emph{The
  International Journal of Robotics Research}, vol.~21, no.~8, pp. 679--699,
  2002. [Online]. Available: \url{https://doi.org/10.1177/027836402761412430}
\BIBentrySTDinterwordspacing

\bibitem{emalis}
\BIBentryALTinterwordspacing
E.~Malis, G.~Chesi, and R.~Cipolla, ``212d visual servoing with respect to
  planar contours having complex and unknown shapes,'' \emph{The International
  Journal of Robotics Research}, vol.~22, no. 10-11, pp. 841--853, 2003.
  [Online]. Available: \url{https://doi.org/10.1177/027836490302210004}
\BIBentrySTDinterwordspacing

\bibitem{Hu_2020_CVPR}
Y.~Hu, P.~Fua, W.~Wang, and M.~Salzmann, ``Single-stage 6d object pose
  estimation,'' in \emph{Proceedings of the IEEE/CVF Conference on Computer
  Vision and Pattern Recognition (CVPR)}, June 2020.

\bibitem{Song_2020_CVPR}
C.~Song, J.~Song, and Q.~Huang, ``Hybridpose: 6d object pose estimation under
  hybrid representations,'' in \emph{Proceedings of the IEEE/CVF Conference on
  Computer Vision and Pattern Recognition (CVPR)}, June 2020.

\bibitem{Yan_Tao_Xu_2022}
S.~Yan, X.~Tao, and D.~Xu, ``Image-based visual servoing system for components
  alignment using point and line features,'' \emph{IEEE Transactions on
  Instrumentation and Measurement}, vol.~71, p. 1–11, 2022.

\bibitem{Al-Shanoon_Lang_2022}
A.~Al-Shanoon and H.~Lang, ``Robotic manipulation based on 3-d visual servoing
  and deep neural networks,'' \emph{Robotics and Autonomous Systems}, vol. 152,
  p. 104041, 2022, citation Key: ALSHANOON2022104041.

\bibitem{pmlr-v155-katara21a}
\BIBentryALTinterwordspacing
P.~Katara, H.~YVS, H.~Pandya, A.~Gupta, A.~Sanchawala, G.~Kumar, B.~Bhowmick,
  and M.~Krishna, ``Deepmpcvs: Deep model predictive control for visual
  servoing,'' in \emph{Proceedings of the 2020 Conference on Robot Learning},
  ser. Proceedings of Machine Learning Research, J.~Kober, F.~Ramos, and
  C.~Tomlin, Eds., vol. 155.\hskip 1em plus 0.5em minus 0.4em\relax PMLR,
  16--18 Nov 2021, pp. 2006--2015. [Online]. Available:
  \url{https://proceedings.mlr.press/v155/katara21a.html}
\BIBentrySTDinterwordspacing

\bibitem{Hutchinson_Hager_Corke_1996}
S.~Hutchinson, G.~Hager, and P.~Corke, ``A tutorial on visual servo control,''
  \emph{IEEE Transactions on Robotics and Automation}, vol.~12, no.~5, p.
  651–670, Oct. 1996.

\bibitem{Costanzo_De_Maria_Natale_Russo_2024}
M.~Costanzo, G.~De~Maria, C.~Natale, and A.~Russo, ``Modeling and control of
  sampled-data image-based visual servoing with three-dimensional features,''
  \emph{IEEE Transactions on Control Systems Technology}, vol.~32, no.~1, p.
  31–46, Jan. 2024.

\bibitem{Luo_Chen_Quan_Zhang_Liu_2020}
B.~Luo, H.~Chen, F.~Quan, S.~Zhang, and Y.~Liu,
  ``\BIBforeignlanguage{en}{Natural feature-based visual servoing for grasping
  target with an aerial manipulator},'' \emph{\BIBforeignlanguage{en}{Journal
  of Bionic Engineering}}, vol.~17, no.~2, p. 215–228, Mar. 2020.

\bibitem{Bateux_Marchand_Leitner_Chaumette_Corke_2017}
\BIBentryALTinterwordspacing
Q.~Bateux, E.~Marchand, J.~Leitner, F.~Chaumette, and P.~Corke, ``Visual
  servoing from deep neural networks,'' no. arXiv:1705.08940, June 2017,
  arXiv:1705.08940 [cs]. [Online]. Available:
  \url{http://arxiv.org/abs/1705.08940}
\BIBentrySTDinterwordspacing

\bibitem{Wu_Jin_Liu_Yu_Yang_2022}
J.~Wu, Z.~Jin, A.~Liu, L.~Yu, and F.~Yang, ``\BIBforeignlanguage{en}{A survey
  of learning-based control of robotic visual servoing systems},''
  \emph{\BIBforeignlanguage{en}{Journal of the Franklin Institute}}, vol. 359,
  no.~1, p. 556–577, Jan. 2022.

\bibitem{Tokuda_Arai_Kosuge_2021}
F.~Tokuda, S.~Arai, and K.~Kosuge, ``Convolutional neural network-based visual
  servoing for eye-to-hand manipulator,'' \emph{IEEE Access}, vol.~9, p.
  91820–91835, 2021.

\bibitem{Puang_Peng_Tee_Jing_2020}
\BIBentryALTinterwordspacing
E.~Y. Puang, K.~Peng~Tee, and W.~Jing, ``Kovis: Keypoint-based visual servoing
  with zero-shot sim-to-real transfer for robotics manipulation,'' in
  \emph{2020 IEEE/RSJ International Conference on Intelligent Robots and
  Systems (IROS)}.\hskip 1em plus 0.5em minus 0.4em\relax Las Vegas, NV, USA:
  IEEE, Oct. 2020, p. 7527–7533. [Online]. Available:
  \url{https://ieeexplore.ieee.org/document/9341370/}
\BIBentrySTDinterwordspacing

\bibitem{Lu_Chen_Lee_Hsu_2023}
\BIBentryALTinterwordspacing
B.-S. Lu, T.-I. Chen, H.-Y. Lee, and W.~H. Hsu, ``Cfvs: Coarse-to-fine visual
  servoing for 6-dof object-agnostic peg-in-hole assembly,'' in \emph{2023 IEEE
  International Conference on Robotics and Automation (ICRA)}.\hskip 1em plus
  0.5em minus 0.4em\relax London, United Kingdom: IEEE, May 2023, p.
  12402–12408. [Online]. Available:
  \url{https://ieeexplore.ieee.org/document/10160525/}
\BIBentrySTDinterwordspacing

\bibitem{Biagiotti_Melchiorri_2008}
L.~Biagiotti and C.~Melchiorri, \emph{\BIBforeignlanguage{eng}{Trajectory
  planning for automatic machines and robots}}.\hskip 1em plus 0.5em minus
  0.4em\relax Berlin Heidelberg: Springer, 2008.

\bibitem{ROS}
M.~Quigley, K.~Conley, B.~Gerkey, J.~Faust, T.~Foote, J.~Leibs, R.~Wheeler, and
  A.~Ng, ``Ros: an open-source robot operating system,'' vol.~3, 01 2009.

\bibitem{Jocher_Ultralytics_YOLO_2023}
\BIBentryALTinterwordspacing
G.~Jocher, A.~Chaurasia, and J.~Qiu, ``{Ultralytics YOLO},'' Jan. 2023.
  [Online]. Available: \url{https://github.com/ultralytics/ultralytics}
\BIBentrySTDinterwordspacing

\bibitem{Ma_Liu_Zhang_Xu_Zhang_Wu_2020}
Y.~Ma, X.~Liu, J.~Zhang, D.~Xu, D.~Zhang, and W.~Wu,
  ``\BIBforeignlanguage{en}{Robotic grasping and alignment for small size
  components assembly based on visual servoing},''
  \emph{\BIBforeignlanguage{en}{The International Journal of Advanced
  Manufacturing Technology}}, vol. 106, no. 11–12, p. 4827–4843, Feb. 2020.

\end{thebibliography}

\end{document}